\documentclass[12pt]{article}
\usepackage{fullpage}
\usepackage{graphicx}
\usepackage{float}
\usepackage{amsmath}
\usepackage{amssymb}
\usepackage{amsthm}
\usepackage{multirow}

\usepackage{hyperref}
\usepackage{booktabs}
\usepackage{url}
\usepackage{threeparttable}
\usepackage[hmargin=1.0in,vmargin=1.0in]{geometry}
\usepackage{authblk}
\usepackage{pdfpages}
\usepackage{color}
\usepackage{setspace}
\usepackage{lineno}
\usepackage{color}
\usepackage{comment}
\usepackage{makecell}
\usepackage{tabularx}
\usepackage{natbib}
\usepackage{threeparttable}
\usepackage{makecell}
\usepackage{booktabs} 
\usepackage{caption} 
\usepackage{algpseudocode}
\usepackage{amsmath}

\begin{document}
\title{Automated Segmentation and Tracking of Group Housed Pigs Using Foundation Models}

\author[1]{Ye Bi}
\author[1,2]{Bimala Acharya}
\author[1]{David Rosero}
\author[1, *]{Juan Steibel}
\affil[1]{Department of Animal Science, Iowa State University, Ames, IA, 50010, USA}
\affil[2]{Interdepartmental Bioinformatics and Computational Biology, Iowa State University, Ames, IA, 50010, USA}

\date{}

\maketitle

\noindent
Running title: Foundation Model–Based Pig Tracking\\


\noindent 
$^{*}$ Corresponding author \\ 

\noindent
ORCID: 0000-0001-7871-5856 (YB), 0000-0002-4679-3443 (BA), 0009-0003-3131-5275 (DR), 0000-0003-1725-2389 (JS)\\ 
 
\noindent
Email addresses: yebi@iastate.edu (YB), bacharya@iastate.edu (BA), dsrosero@iastate.edu (DR), jsteibel@iastate.edu (JS) \\

\newpage
\doublespacing

\section*{Abstract} 
Foundation models (FM) are reshaping computer vision by reducing reliance on task-specific supervised learning and leveraging general visual representations learned at scale. In precision livestock farming, most pipelines remain dominated by supervised learning models that require extensive labeled data, repeated retraining, and farm-specific tuning. This study presents an FM-centered workflow for automated monitoring of group-housed nursery pigs, in which pretrained vision-language FM serve as general visual backbones and farm-specific adaptation is achieved through modular post-processing. Grounding-DINO was first applied to 1,418 annotated images to establish a baseline detection performance. While detection accuracy was high under daytime conditions, performance degraded under night-vision and heavy occlusion, motivating the integration of temporal tracking logic. Building on these detections, short-term video segmentation with Grounded-SAM2 was evaluated on 550 one-minute video clips; after post-processing, over 80\% of 4,927 active tracks were fully correct, with most remaining errors arising from inaccurate masks or duplicated labels. To support identity consistency over an extended time, we further developed a long-term tracking pipeline integrating initialization, tracking, matching, mask refinement, re-identification, and post-hoc quality control. This system was evaluated on a continuous 132-minute video and maintained stable identities throughout. On 132 uniformly sampled ground-truth frames, the system achieved a mean region similarity (J) of 0.83, contour accuracy (F) of 0.92, J\&F of 0.87, MOTA of 0.99, and MOTP of 90.7\%, with no identity switches. Overall, this work demonstrates how FM prior knowledge can be combined with lightweight, task-specific logic to enable scalable, label-efficient, and long-duration monitoring in pig production.

\newpage
\section{Introduction} 
Pork is the world’s most consumed meat, with over one billion pigs farmed annually and producing more than 116.45 million tons of meat (USDA 2024). Modern pig production systems confine large populations of animals, with by relatively few caregivers overseeing them, making effective health and welfare monitoring increasingly challenging~\citep{berckmans2017general}. This challenge motivates the development of precision livestock farming (PLF), which utilizes sensors and machine learning to enable continuous, automated monitoring of individual animals~\citep{norton2017developing}. Conventional surveillance based on human observation is labor-intensive and often unable to detect the subtle or rapidly progressing signs of diseases~\citep{matthews2017automated, norton2019precision, fernandes2020image}. By contrast, computer vision (CV) based monitoring systems offer a non-invasive, low-stress, and cost-effective solution that enables continuous tracking of individual animals~\citep{wurtz2019recording, neethirajan2021digital}. Vision-based pig recognition is particularly critical because it establishes a ``digital representation” for each pig, enabling them to be connected with key variables relevant to management and productivity~\citep{norton2019precision}.

In recent years, deep learning (DL)-based approaches, particularly supervised learning models (SLM), have attracted increasing attention across a wide range of PLF research works. These include body measurement such as body weight~\citep{bi2025industry}, body condition score~\citep{shuai2020research}, pig posture recognition (e.g., sternal lying, lateral lying, and standing)~\citep{bhujel2021deep}, and behavior monitoring tasks such as feeding~\citep{yang2018feeding, chen2020recognition}, drinking~\citep{chen2020classification}, aggression~\citep{han2023evaluation}, and injurious behaviors like mounting~\citep{yang2021pig} and tail biting~\citep{liu2020computer}. To support these diverse applications, widely adopted neural network architectures include YOLO, ResNet-50, VGG-16, DenseNet, Faster R-CNN, Mask R-CNN, and long short-term memory~\citep{li2022barriers, bi2023depth, rohan2024application}. A common workflow involves using DL models pretrained on large-scale image datasets such as ImageNet~\citep{deng2009imagenet} or Microsoft COCO~\citep{lin2014microsoft}, followed by supervised fine-tuning for specific tasks. However, this method limits fast and scalable deployment because it relies on large, labor-intensive labeled datasets, it demands specialized expertise for model design and parameter tuning, it requires costly retraining when applied to new farms or conditions, and often fails to generalize robustly to unseen environments~\citep{kamilaris2018deep, alameer2020automated, bhujel2025systematic}.

To overcome these challenges, recent advances in large pre-trained foundation models (FM) offer a promising direction. FM are generally trained through self-supervised learning~\citep{liu2021self}, on extensive datasets drawn from diverse domains, and can be efficiently adapted to new tasks with limited labeled data and minimal fine-tuning~\citep{bommasani2021opportunities, awais2025foundation}. FM have demonstrated remarkable versatility across modalities, including language, vision, vision–language integration, and generative modeling.

Within PLF, there is growing interest in leveraging FM's generalization capabilities to support a diverse range of visual tasks. A major emerging direction is utilizing FM as annotation engines for expanding large-scale datasets. For instance, the combination of Grounding-DINO and HQ-SAM~\citep{ke2023segment} has enabled automatic generation of segmentation masks for pigs, sheep, and cattle at a large scale, substantially reducing the manual effort required for pixel-level annotation~\citep{wu2024accelerated}. From a data augmentation perspective, the FTO-SORT pipeline integrates SAM and the image-inpainting model, LaMa, to generate realistic synthetic pig images by copy-pasting pig contours into diverse backgrounds, thereby enhancing dataset diversity for downstream training~\citep{yu2025fto}. 

Beyond dataset expansion, FM have also been adopted as core backbones for visual tasks, including detection, segmentation, and tracking~\citep{awais2025foundation}. For example, SAM combined with 3D imaging and ResNet architectures has been applied for non-invasive pig weight estimation~\citep{bharadwaj2025non}, while SAM coupled with a random forest framework has enabled poultry weight prediction from thermal images~\citep{yang2024innovative}. In animal tracking, SAM has been integrated with a supervised detector (YOLO~\citep{sirisha2023statistical}) and tracker (ByteTracker~\citep{zhang2021bytetrack}) for real-time monitoring of cage-free hens, illustrating the adaptability of FM to different sensor modalities and species~\citep{yang2024innovative}.

FM have also shown great promise in behavior analysis, a key component of precision animal monitoring. AnimalFormer, for example, unifies Grounding-DINO, HQ-SAM, and ViTPose~\citep{xu2022vitpose} to extract masks and key points for sheep, enabling unsupervised clustering of postures and movements~\citep{dussert2025zero}. Likewise, FM such as OWLv2~\citep{minderer2023scaling}, SAMURAI~\citep{yang2024samurai}, and DINOv2~\citep{oquab2023dinov2} have been combined to extract segmentation-based features for supervised pig behavior analyses~\citep{yang2025computer}. Yet, these existing frameworks remain limited to short video segments and lack mechanisms for maintaining consistent identities across extended recordings, such as the multi-hour videos considered in this study. Collectively, these studies demonstrate the versatility and unifying capability of FM across key visual tasks, signaling a paradigm shift toward foundation-model-driven PLF systems.

Despite these advances, no existing study has utilized FM alone, without any supervised components, for long-term video monitoring of group-housed pigs. To fill this gap, we propose a fully FM-based pipeline built upon Grounded-SAM2-based, which jointly performs pig detection, instance segmentation, and multi-object tracking across both static images and long video sequences. The objectives of this study are threefold: (1) to evaluate the performance of FM for group-housed pig monitoring across different temporal scales; (2) to develop a unified FM pipeline for detection, segmentation, and long-term tracking; (3) to identify and discuss the practical limitations and future opportunities of applying such models in real farm environments.

This study demonstrates the feasibility of foundation-model-driven, label-free, and scalable long-term video monitoring for group-housed pigs, providing a critical step toward autonomous PLF.

\section{Conceptual Design}
To address these motivations and objectives, our approach rethinks the conventional workflow of animal monitoring by removing all dependencies on supervised models and manual annotations, introducing a paradigm shift toward an FM-based framework. The FM eliminates the need for ground-truth labeling and task-specific model pre-training before model inference. Labels are only applied post-hoc for evaluation, thereby significantly reducing manual effort and improving scalability. 

\begin{figure}[H]
    \centering
    \includegraphics[width=1\textwidth]{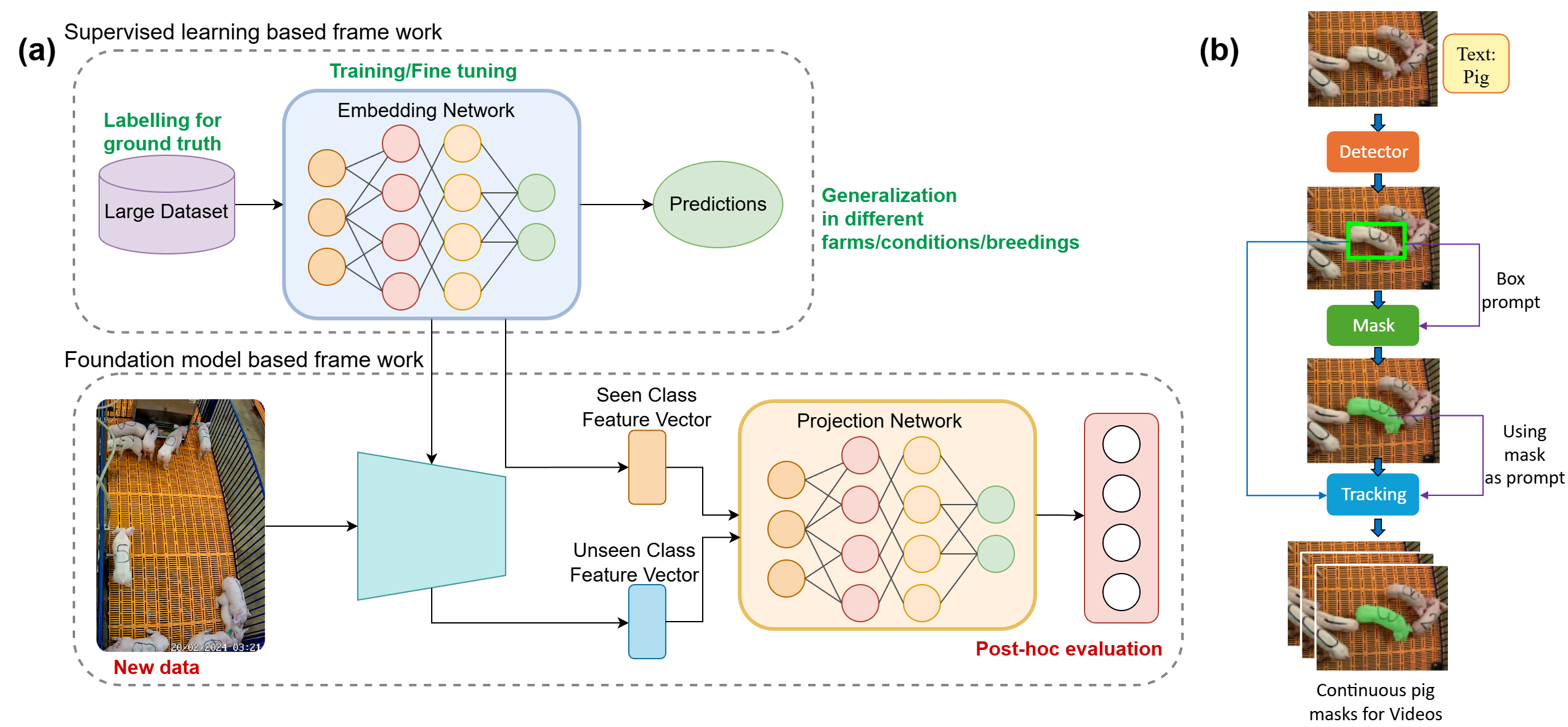}
    \caption{(a) Conceptual comparison between conventional supervised learning and a foundation-model (FM)–based framework. (b) FM-based pipeline for pig video segmentation and tracking, including text-prompted detection, mask generation, and mask-based tracking across video frames.}
\label{fig:gsam2_arc}
\end{figure}

Building on this conceptual design, we implement a vision-language foundation model pipeline upon Grounded-SAM2 to evaluate the effectiveness of FM for pig detection, video segmentation, and tracking \citep{ren2024groundedsamassemblingopenworld}. This framework combines object detection and prompt-based segmentation with temporal tracking, as illustrated in Fig.~\ref{fig:gsam2_arc} (a).

\textbf{Detection}: The detection component utilizes Grounding-DINO, a vision-language foundation model capable of detecting arbitrary objects based on human-provided text prompts, e.g., category names or referring expressions~\citep{liu2024groundingdinomarryingdino}. Grounding-DINO employs a Transformer-based architecture that performs joint reasoning over image and text inputs to generate bounding box predictions (Fig.~\ref{fig:gsam2_arc}(b)).

\textbf{Segmentation}: Following detection, the predicted bounding boxes are passed to SAM2 to produce high-resolution, pixel-level segmentation masks~\citep{ravi2024sam}. SAM2 leverages a prompt-based mask decoder to convert the box-level prompts into instance masks with fine spatial detail (Fig.~\ref{fig:gsam2_arc}(c)).

\textbf{Tracking}: To extend segmentation across time, the tracking component of SAM2 is employed, which propagates object masks across video frames using a memory attention mechanism, enabling stable and consistent segmentation throughout the sequence without requiring manual annotations for each frame (Fig.~\ref{fig:gsam2_arc}(c)).

With this framework design, individual pig bodies can be detected and segmented in short video clips simply by providing the text prompt ‘pig’ and the pig videos, without requiring model fine-tuning or manual data labeling. However, the original Grounded-SAM2 model faces limitations in addressing key challenges associated with long-term pig tracking, such as maintaining identity consistency, severe occlusions, and re-identifying animals that temporarily disappear and later reappear in the scene. We addressed these challenges by extending the original Grounded-SAM2 with a modular pipeline that links the last frame of one video clip to the first frame of the next, ensuring temporal continuity and identity consistency across clips, as described in the following sections.

\section{Materials and Methods}
\subsection{Animals and Data Collection Design}
The data used in this experiment were collected as part of a larger trial performed at Iowa State University. The animals used in the trial were crossbred white pigs (L42~$\times$~L337, PIC), with an initial body weight of \(5.86 \pm 1.01\)~kg and an average age of 18 days at the start of the experiment. The pigs were housed in groups of ten animals per pen, with each pen providing a standard nursery environment. Each pig was individually marked with a unique painted number on its back. The animals were fed daily under an \textit{ad libitum} feeding regime with a common commercial diet and had continuous access to clean drinking water.

\subsection{Video Acquisition Setup}
\begin{figure}[H]
    \centering
    \includegraphics[width=1\textwidth]{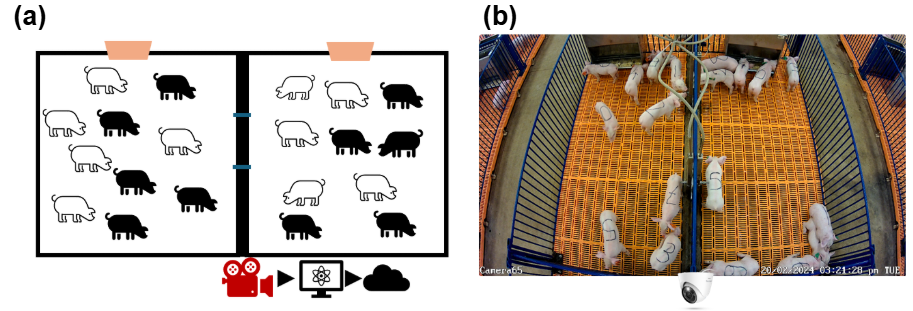}
    \caption{Experimental setting on nursery pig farm. (a-b) Schematic and real views of the camera-pen arrangement.}
\label{fig:exp_set}
\end{figure}

Video data were collected using six wireless PoE security cameras (Reolink RLC-833A), each offering a horizontal field of view of 94$^\circ$ and a vertical field of view of 50$^\circ$. The cameras were mounted in fixed locations between adjacent pens, positioned to provide a clear high-angle view of the pigs (see Fig.~\ref{fig:exp_set} for layout). Continuous video was recorded over four days post-weaning, with RGB footage during daylight and infrared gray-scale footage at night under low-light conditions. In total, video recordings were obtained from 12 pens, covering 120 pigs. Individual video clips had durations of approximately 22 minutes and were captured continuously throughout the four-day observation period. Each video was recorded at a resolution of 2560~$\times$~1440 pixels (width~$\times$~height) and a frame rate of 10 fps. The recorded videos were automatically uploaded from the farm to a secure cloud server via a stable wired internet connection. In total, the raw video dataset comprised 0.97 TB of footage.

To facilitate the application of foundation vision-language models, we selected three task-specific subsets from these collected data: (1) animal detection (section~\ref{sec:animal_detection}), (2) short-term video segmentation (section~\ref{sec:short-seg}), and (3) long-term video segmentation (section~\ref{sec:long-seg}).

\subsection{Animal Detection}
\label{sec:animal_detection}
We used a publicly available dataset of 1,418 annotated frames extracted from the collected videos near the feeder area, with pigs labeled using bounding boxes if at least 50\% of their body was visible and they were inside the pen. The dataset is available on OSF (\url{https://osf.io/pxts8/}) and was previously described at USPLF 2025~\citep{steibel2025early}.

For this object detection task, we applied Grounding-DINO with the text prompt ``\textit{pig}” throughout all experiments. Among the available pretrained variants, we selected the Tiny model (Swin-Tiny backbone, 172M parameters) for its computational efficiency, which was well-suited for our single-class, multi-instance dataset. 

To optimize detection performance, we tuned two key Grounding-DINO hyperparameters: \texttt{BOX\_THRESHOLD} (confidence filtering) and \texttt{TEXT\_THRESHOLD} (text-visual alignment), varying both from 0.01 to 0.60. A predicted box was considered correct if its Intersection over Union (IoU, equation~\ref{eq:iou}) with a ground truth box exceeded 0.5. We evaluated detection results using the following metrics:
\begin{align}
\text{IoU}(b_{\text{pred}}, b_{\text{gt}}) &= \frac{\text{area}(b_{\text{pred}} \cap b_{\text{gt}})}{\text{area}(b_{\text{pred}} \cup b_{\text{gt}})} \label{eq:iou} \\
\text{Precision} &= \frac{TP}{TP + FP} \label{eq:precision} \\
\text{Recall} &= \frac{TP}{TP + FN} \label{eq:recall} \\
\text{F1 Score} &= \frac{2 \cdot \text{Precision} \cdot \text{Recall}}{\text{Precision} + \text{Recall}} \label{eq:f1}
\end{align}

Here, $TP$ denotes true positives (predicted boxes with IoU $\geq 0.5$), $FP$ false positives (non-matching or spurious detections), and $FN$ false negatives (missed ground truth boxes). 

\subsubsection{Post hoc evaluation}
\label{sec:post_hoc4detection}
We selected the model configuration yielding the highest F1 score for post hoc evaluation. Three conditions were examined: (1) whether a bounding box was located inside or outside the pen, (2) whether the image was captured under day or night vision, and (3) whether the bounding box covered more or less than 50\% of a pig’s body.

\subsubsection{Short-term video segmentation}
\label{sec:short-seg}
After evaluating Grounding DINO, we conducted short-term tracking during daytime hours (06:00–17:00) on two out of the four days immediately post-mixing. Five 1-minute videos were extracted every 64 minutes from 6 cameras, resulting in 550 video clips for segmentation. All frames from the same camera were then cropped to the right-side region and resized to a fixed width of 480 pixels, preserving the aspect ratio.

For each cropped video clip, we applied Grounded-SAM2 to segment and track individual pigs over time. Using the prompt ``\textit{pig}", detections were generated by Grounding-DINO with both text and box thresholds set to 0.24. If more than 15 pigs were detected, only the detections with the top 10 confidence scores were retained to ensure quality. The selected boxes were processed as box prompts by SAM2 for video segmentation. Among SAM2's four pretrained variants (Tiny, Small, Base+, Large), we initially tested the Tiny model for its balance of speed and segmentation quality.

The final outputs included per-frame segmentation data in JSON format and annotated MP4 videos, each include enhanced images with each pig body highlighted in a different color, this facilitates the post hoc evaluation by human observers. Finally, we generated 4,927 auto-annotated videos for two days, covering 2.96 million frames. 

\subsubsection{Post hoc evaluation}
After applying FMs, a human evaluator reviewed each annotated video to flag tracking errors (e.g., incorrect mask, duplicated label, ID switch, or lost track) and video-level challenges (e.g., pigs outside the pen or stacking) using a structured evaluation sheet. The definitions of these error types and challenges are summarized in Table~\ref{tab:track_error}. We quantified pig stacking dynamics and tracking reliability by computing the hourly counts of stacked versus unstacked pigs and the corresponding unstacked ratio. In addition, by manually categorizing four types of tracking errors from the auto-annotated videos, the percentage of each error was summarized in active pig tracks for two days. Twenty video clips and their corresponding annotated JSON files, including both standard and challenging tracking scenarios, have been uploaded to OSF (\url{https://osf.io/63ndh/}).

\subsubsection{Tracking model improvement}
After red-flagging these issues, we employed the following strategies to enhance the FM's performance for short-term video segmentation.

To remove tracking errors caused by mistakenly detecting pigs outside the pen, we applied spatial filtering using known pen boundaries. Only detections with at least 40\% of their area inside the pen were retained.

To handle duplicated labels, we applied mask-level non-maximum suppression. Masks with more than 8\% overlaps were compared, and the one with lower detection confidence was discarded. This ensured only the highest-confidence mask was retained among overlapping detections.

To minimize ID switch and lost track errors, we employed two strategies. First, we replaced the Tiny model of SAM2 with the Large model, which provides greater representational capacity for complex scenes. As we noticed, ID switches often occurred when pigs partially overlapped with each other, leading to the tracks merging or switching. The second strategy to correct this, as illustrated in Fig.~\ref{fig:rev_sam2}, is that we identified the first frame where the ID switch occurred in this section by watching the annotated video, then applied Grounding DINO to detect pigs frame by frame until a valid anchor frame with exactly ten detections was found, assuming that the frame served as a better track starting point. From this point, we applied SAM2 tracking bidirectionally: backward to the error frame and forward to the end of the video. The corrected segment replaced the erroneous one. If no valid anchor was found, the error segment was excluded to prevent error propagation.

To address incorrect masks, we identified that many of these tracking errors originated from unreliable detections provided by Grounding-DINO, particularly in scenes where pigs were in close contact or heavily occluded. In such cases, accurate segmentation cannot be achieved solely through detection in a single frame. We describe our strategy for handling these situations in Section~\ref{sec:long-seg}, where we show that as long as tracking can be reinitialized from a clean frame, the results remain unaffected by incorrect masks. Additionally, based on the review of annotation videos, sometimes incorrect masks were introduced during the tracking process itself. These were corrected through a dedicated mask post-editing procedure applied after SAM2 tracking, as detailed in the same section. 

\subsection{Long-term video segmentation}
\label{sec:long-seg}
We developed a zero-shot long-term pig video segmentation pipeline based on modified Grounded-SAM2, capable of maintaining consistent identities and pig segmentation across 1-minute video segments recorded in a medium-density pig nursery. As shown in Fig.~\ref{fig:flowchart}, we split each component of the flowchart into the following modules: initializer, tracker, matcher, mask refinement, re-identification (ReID), and post-segment-quality control (post-QC). The initializer module selected a clean reference frame using detection and segmentation to establish initial pig identities. The tracker module propagated the masks frame by frame within each 1-minute clip using the SAM2 video predictor. The matcher module linked pig identities across consecutive clips to construct long-range trajectories. The mask refinement module removed duplicated or erroneous segmentation. The re-identification module restored missing or lost identities. Finally, the post quality control module flagged and filtered out unreliable results, ensuring overall consistency. The main goal was to extend short-term video segmentation into a long-term framework by accurately linking and maintaining consistent pig identities across consecutive video segments.

We selected six 22-minute video recordings, totaling approximately two hours, from a single camera between 06:25 and 08:34 in the early morning. This period captures a typical behavioral transition, during which most pigs are initially active, such as standing and feeding, before gradually lying down to rest approximately one hour later. Each video was segmented into 1-minute clips, and only the right-side region of the pen was retained for evaluation, following the same procedure described in Section~\ref{sec:short-seg} and illustrated in Fig.~\ref{fig:exp_set}.

\subsubsection{Initializer}

The initializer module identified a good reference frame to initialize pig identities for long-term video segmentation. It scanned every 10th frame in the input sequence and applied the full detection and segmentation pipeline, with Grounding-DINO generating bounding boxes and SAM2 producing the corresponding masks, as shown in Fig.~\ref{fig:gsam2_arc}. To ensure spatial precision and eliminate ambiguity, the results were filtered using predefined pen boundaries and a mask-level non-maximum suppression strategy, as described in Section~\ref{sec:short-seg}. A frame was accepted as the reference if it contained exactly 10 high-confidence, non-overlapping pigs fully within the pen when pigs were more active. In later clips over 2 hrs, when pig activity decreased, and occlusion or piling occurred; some pigs became invisible for extended periods. In such cases, we accepted frames with fewer than 10 visible pigs by applying the Matcher module (Section~\ref{sec:matcher}), which adaptively adjusted the expected number of pigs based on their location and movement. This strategy ensured a reliable starting point for subsequent tracking and maintaining identities over time.

\subsubsection{Tracker}
\begin{figure}[H]
    \centering
    \includegraphics[width=1\textwidth]{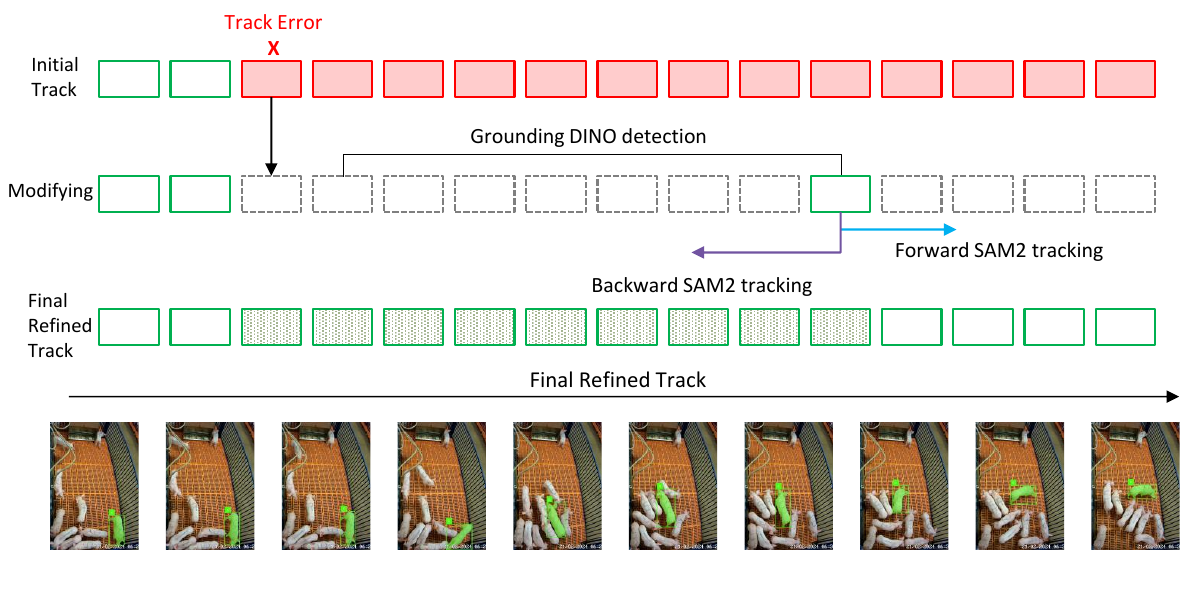}
    \caption{Bidirectional SAM2 propagation diagram. After identifying the first erroneous frame (red cross), GroundingDINO was used to find a valid anchor frame (green box), and SAM2 tracking was applied backward (purple arrow) and forward (blue arrow) to replace the erroneous segment with a corrected track.}
\label{fig:rev_sam2}
\end{figure}

\begin{figure}[H]
    \centering
    \includegraphics[width=0.8\textwidth]{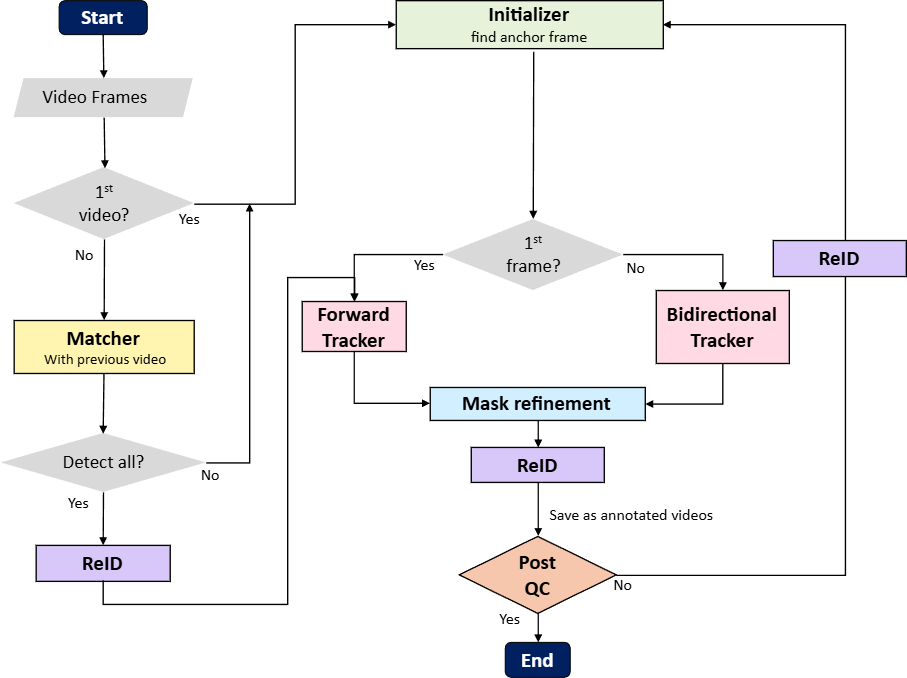}
    \caption{Long-term video segmentation workflow. The pipeline integrates six modules to extend short-term segmentation into long-term tracking across consecutive video clips: initializer (identity assignment from a clean reference frame), tracker (mask propagation within clips), matcher (cross-clip identity linking), mask refinement (removing duplicate or erroneous masks), re-identification (restoring missing IDs), and post-segmentation quality control (flagging unreliable results).}
\label{fig:flowchart}
\end{figure}

The filtered masks from the selected reference frame were used to initialize the tracker (SAM2) module for propagating pig masks across time. If the reference frame was the first in the video clip, forward-only propagation was performed. Otherwise, bi-directional inference was applied, with forward tracking toward the end of the clip and backward tracking toward the beginning. The two tracks were then merged to produce a temporally consistent and identity-preserving segmentation sequence, as illustrated in “Final Refined Track” in Fig.~\ref{fig:rev_sam2}. This bidirectional strategy marked a key improvement over the original SAM2, enabling robust tracking even under occlusion and pig stacking.

The tracker module was also employed to recover from tracking errors. As shown in Fig.\ref{fig:flowchart} and Fig.~\ref{fig:rev_sam2}, the post quality control (post QC) module automatically identifies error frames. Section~\ref{sec:short-seg} describes the overall error-handling strategy. The key difference in this section is that error frames were flagged automatically by the post-QC module, rather than being manually flagged. Once re-tracking was completed, the new video segments were aligned with previous ones using the ReID module.

\subsubsection{Mask refinement}
To improve segmentation accuracy and temporal coherence, we applied a rule-based mask refinement procedure guided by geometric constraints and inter-frame continuity. The process involved spatial filtering within the pen, followed by blob analysis and centroid tracking across frames. Candidate regions were evaluated based on area, distance from the primary blob, and proximity to the previous centroid. Invalid regions were discarded, and centroids of valid blobs were updated to guide refinement over time. The full logic is summarized in Supplementary Algorithm~1: Mask Refinement Procedure.

\subsubsection{Matcher}
\label{sec:matcher}
The Matcher module was responsible for matching pig identities between consecutive clips by selecting the best frame from the previous clip, typically one of the last ten frames, containing 10 sufficiently large and non-overlapping masks. Segmentation masks from this frame were transferred to the new clip, and bounding boxes were used to prompt SAM2 to generate masks in the first frame. If a close temporal match was found, direct SAM2 propagation followed; otherwise, the Matcher module iteratively searched earlier frames until a valid initialization was achieved, and the ReID module was applied to align identities.

In cases where fewer than 10 pigs were consistently visible in the previous clip, a fallback strategy was applied: Grounding DINO was first used to detect all pigs in the new clip. If a clean frame with 10 confident detections was identified, SAM2 propagated masks through the remainder of the clip, and the ReID module was used to align identities with those from the previous segment. If detection failed to recover 10 pigs, the Matcher module adaptively reduced the expected pig count by excluding consistently missing individuals based on their recent visibility. This approach ensured that only reliable instances were propagated forward, enabling robust identity continuity despite occlusion, inactivity, or behavioral clustering.

\subsubsection{Re-identification}
\label{sec:reid}
Although identification (ReID) was not the primary focus of this work, we incorporated a lightweight, feature- and location-aware module to ensure temporal ID consistency across video segments. For each pig mask, the cropped region was contrast-enhanced using gamma correction and histogram equalization, then passed through the SAM2 vision encoder to extract feature vectors. To emulate the temporal memory mechanism of SAM2, we further combined these feature embeddings with centroid proximity and mask overlap into a weighted cost score, and the resulting cost matrix was solved using the Hungarian algorithm to optimally match objects across frames~\citep{kuhn1955hungarian}. This process improved temporal consistency and reduced identity switches, as detailed in Supplementary Algorithm~1.

\subsubsection{Post quality control}
The final tracking results are converted into structured JSON files and annotated MP4 videos by using the same strategy described in Section~\ref{sec:short-seg}. To ensure output quality, we implement a post-quality-control (Post-QC) module to detect and correct tracking errors. Frames with overlapping masks or near-zero area are flagged. This module compared segmentation masks across adjacent frames near red-flag frames to identify inconsistencies such as identity switches or lost tracks. Continuous error sequences are identified and either reprocessed using bidirectional propagation or excluded from the dataset. This module generates tracking performance reports, including masks with huge overlaps and zero areas, and these frames will be evaluated by a human evaluator. This QC stage enforces high-confidence filtering, ensuring the final outputs meet the quality required for scientific and agricultural applications.

\subsubsection{Post hoc evaluation}
To evaluate the accuracy and robustness of our long-term video segmentation and tracking framework, we manually annotated 132 uniformly sampled frames from a 2-hour video at the 200th frame of each minute of recording. Segmentation quality was assessed using two standard metrics: region similarity (Jaccard index, $\mathcal{J}$), which measures the intersection-over-union between predicted and ground-truth masks, and contour accuracy (F-measure, $\mathcal{F}$), which evaluates boundary alignment. We also report the combined $\mathcal{J}\&\mathcal{F}$ mean. For temporal tracking performance, we compute multiple object tracking accuracy (MOTA), which accounts for false positives, missed detections, and identity switches, and multiple object tracking precision (MOTP), which measures localization accuracy based on IoU. In equations~\eqref{eq:jaccard}--\eqref{eq:motp}, $\mathcal{M}_{\text{pred}}$ denotes the predicted segmentation mask and $\mathcal{M}_{\text{gt}}$ denotes the ground truth segmentation mask. $\mathcal{P}_c$ and $\mathcal{R}_c$ represent the contour-based precision and recall.  In the tracking metrics, $FN_t$, $FP_t$, and $IDSW_t$ denote the number of false negatives, false positives, and identity switches at frame $t$, respectively, while $GT_t$ is the number of ground truth objects at that frame. The metric $IoU_{t,i}$ denotes the Intersection over Union between the predicted and ground truth region of object $i$ in frame $t$, and $c_t$ is the number of correctly matched object pairs in frame $t$. 
\begin{align}
\mathcal{J}(\mathcal{M}_{\text{pred}}, \mathcal{M}_{\text{gt}}) &= \frac{|\mathcal{M}_{\text{pred}} \cap \mathcal{M}_{\text{gt}}|}{|\mathcal{M}_{\text{pred}} \cup \mathcal{M}_{\text{gt}}|} \label{eq:jaccard} \\
\mathcal{F} &= \frac{2 \cdot \mathcal{P}_c \cdot \mathcal{R}_c}{\mathcal{P}_c + \mathcal{R}_c} \label{eq:fmeasure} \\
\text{J\&F Mean} &= \frac{1}{2} \left( \overline{\mathcal{J}} + \overline{\mathcal{F}} \right) \label{eq:jfmean} \\
\text{MOTA} &= 1 - \frac{\sum_{t} (FN_t + FP_t + IDSW_t)}{\sum_{t} GT_t} \label{eq:mota} \\
\text{MOTP}_{IoU} &= \frac{\sum_{t,i} IoU_{t,i}}{\sum_t c_t} \label{eq:motp}
\end{align}

\newpage
\section{Results}
\label{sec:results}
We developed a tracking pipeline based on the foundation model, as well as incorporating farm-specific adaptation through modular post-processing, and applied it to video data collected from the group-housed nursery pig farm. Our pipeline can also identify systematic issues and highlight areas that may require targeted annotation and retraining if future model optimization is pursued. The evaluation covered three basic computer vision tasks: animal detection, short-term tracking, and long-term tracking, reflecting increasing levels of difficulty in maintaining accurate pig identities over time.

\subsection{Animal Detection}
\begin{figure}[H]
    \centering
    \includegraphics[width=1\textwidth]{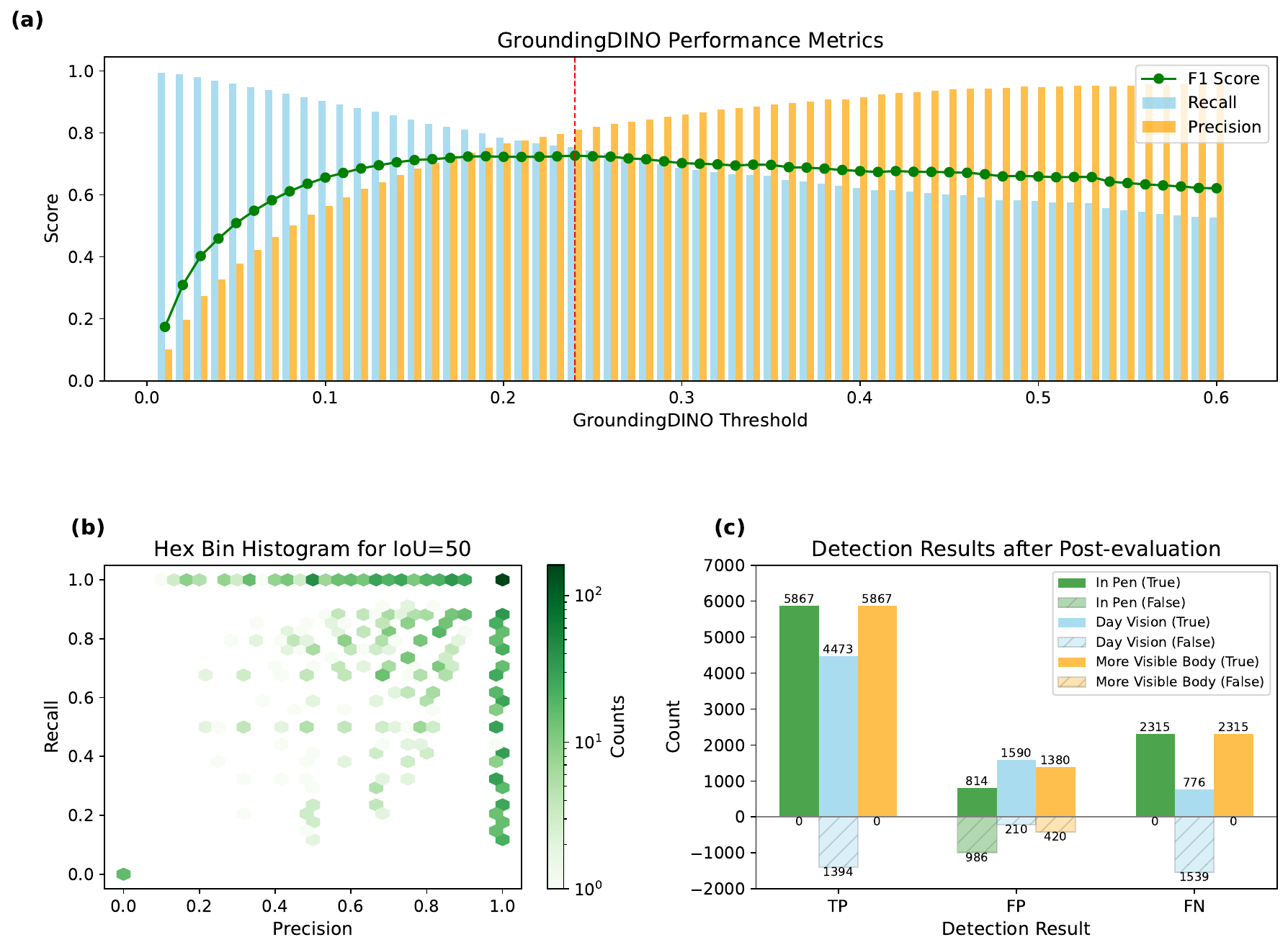}
    \caption{Grounding-DINO performance for pig detection. (a) F1 score, recall, and precision across different Grounding DINO thresholds. (b) Hexbin histogram comparing recall and precision before post-evaluation. (c) Results after post-evaluation, TP, FP, and FN represent true positive, false positive, and false negative, respectively.}
\label{fig:groundingdino_results}
\end{figure}

The first task focused on animal detection in static images. We evaluated Grounding DINO’s detection performance using a ground truth-ready public dataset containing 1,418 images by systematically varying the thresholds for bounding boxes and text prompts. Figure~\ref{fig:groundingdino_results}(a) presents the F1, Recall, and Precision across different threshold values. As expected, increasing the threshold reduces Recall but increases Precision, as higher thresholds lead to fewer detected bounding boxes, but more boxes have overlaps with the ground truth, making them more accurate. The F1 score initially increases and then plateaus. The red dashed line in Fig.\ref{fig:groundingdino_results}(a) marks the highest F1 score (at a threshold of 0.24). 

At this optimal threshold, detection performance was further illustrated in Fig.~\ref{fig:groundingdino_results}(b) as a hexbin plot~\citep{carr1987scatterplot} of precision versus recall, with hexagon intensity representing the number of images. The densest region corresponds to 161 images with both high precision and high recall, indicating strong agreement with the ground truth. Across the dataset, detection yielded 5,867 true positives (TP), 814 false positives (FP), and 2,315 false negatives (FN).  

\begin{figure}[H]
    \centering
    \includegraphics[width=1\textwidth]{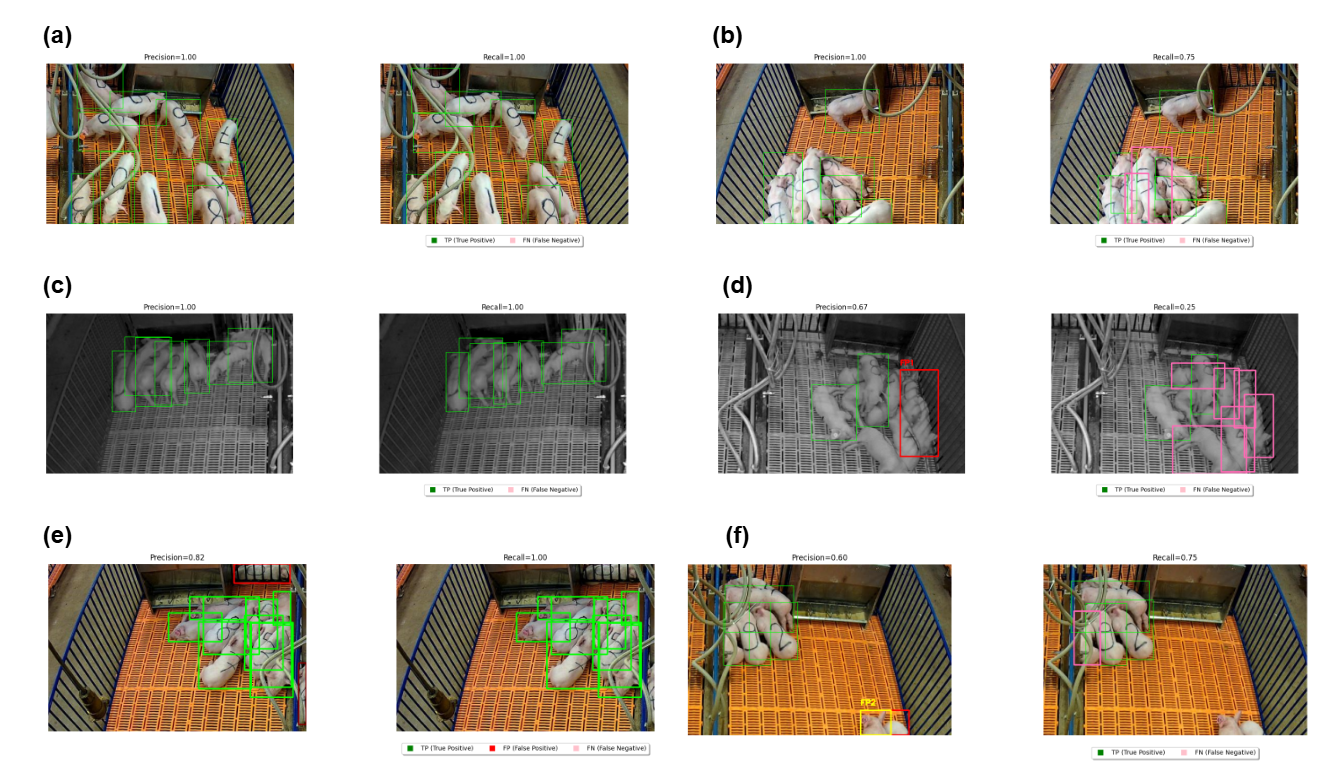}
    \caption{Pig detection examples by raw GroundingDINO under different conditions. In each panel, the left image shows the prediction result, and the right image shows the ground truth. Detection was generally more accurate in daytime color images than in nighttime grayscale images, and when pigs were more separated rather than stacked. (a) Daytime, active pigs. (b) Daytime, inactive pigs. (c) Nighttime, active pigs. (d) Nighttime, inactive pigs. (e) False positive: pig outside the pen. (f) False positive: partial pig detection.}
\label{fig:detection_examples}
\end{figure}

A detailed post-evaluation of detection performance was conducted, as summarized in Fig.~\ref{fig:groundingdino_results}(c). We evaluated detection performance according to the methods described in Section~\ref{sec:post_hoc4detection}. Among all the detected boxes, only 986 were identified as being outside the pen. Most detections within pen boundaries were correct (5867 TP vs. 814 FP). The true positive rate (TPR) under day vision was 85.20\%, higher than the 57.53\% TPR under night vision, indicating that the zero-shot model performed better on RGB daytime images compared to grayscale nighttime images. False negatives were also more frequent under night vision (1,539) than in day vision (776), suggesting that low-light conditions reduce detection reliability. For example, in Fig.~\ref{fig:detection_examples} (a) and (d), the detection was more successful during the daytime compared to nighttime, as pig body contours were clearer in color images than in grayscale night-vision images. Detection performance also improved when pigs were more separated from one another under well-lit conditions (Fig.~\ref{fig:detection_examples} (a,c)), whereas accuracy declined when pigs were stacked closely together (Fig.~\ref{fig:detection_examples} (b,d)).

Many of the FP detections arise from differences between the model’s predictions and our customized ground-truth labeling criteria. For example, the model frequently identified pigs outside the pen, 986 detections versus 814 inside-pen detections. The model also classified partially visible animals as positives, resulting in 420 detections covering less than half of a pig’s body that were not labeled in the ground truth, as shown in  Fig.\ref{fig:groundingdino_results}(c) and Fig.~\ref{fig:detection_examples}(e, f). These discrepancies highlight that some FP cases are not necessarily detection failures but rather reflect customized labeling decisions tailored to this dataset. 

Similarly, several FN cases occurred in challenging scenarios, such as heavy occlusion or partial visibility due to objects or pigs covering the area, where ground truth annotations adhered to strict labeling conventions, as shown in Fig.~\ref{fig:groundingdino_results}(c) and Fig.~\ref{fig:detection_examples}(f). These instances demonstrate that some missed detections are influenced by annotation choices and dataset design, highlighting the inherent difficulty of creating comprehensive ground-truth labels in complex farm environments. 

All the above detection errors may propagate to subsequent segmentation and tracking tasks; however, the advantage of SAM2-based video segmentation is that once an animal becomes fully visible, the mask can be expanded to cover its entire body, mitigating some of the initial detection errors.

\subsection{Short-Term Video Segmentation}
The next task focused on continuous tracking and segmentation in 1-minute videos. After applying Grounded-SAM2, a trained human evaluator reviewed 4,927 auto-annotated videos (2.96 million frames) to document video challenges and tracking errors.

\begin{figure}[H]
    \centering
    \includegraphics[width=0.8\textwidth]{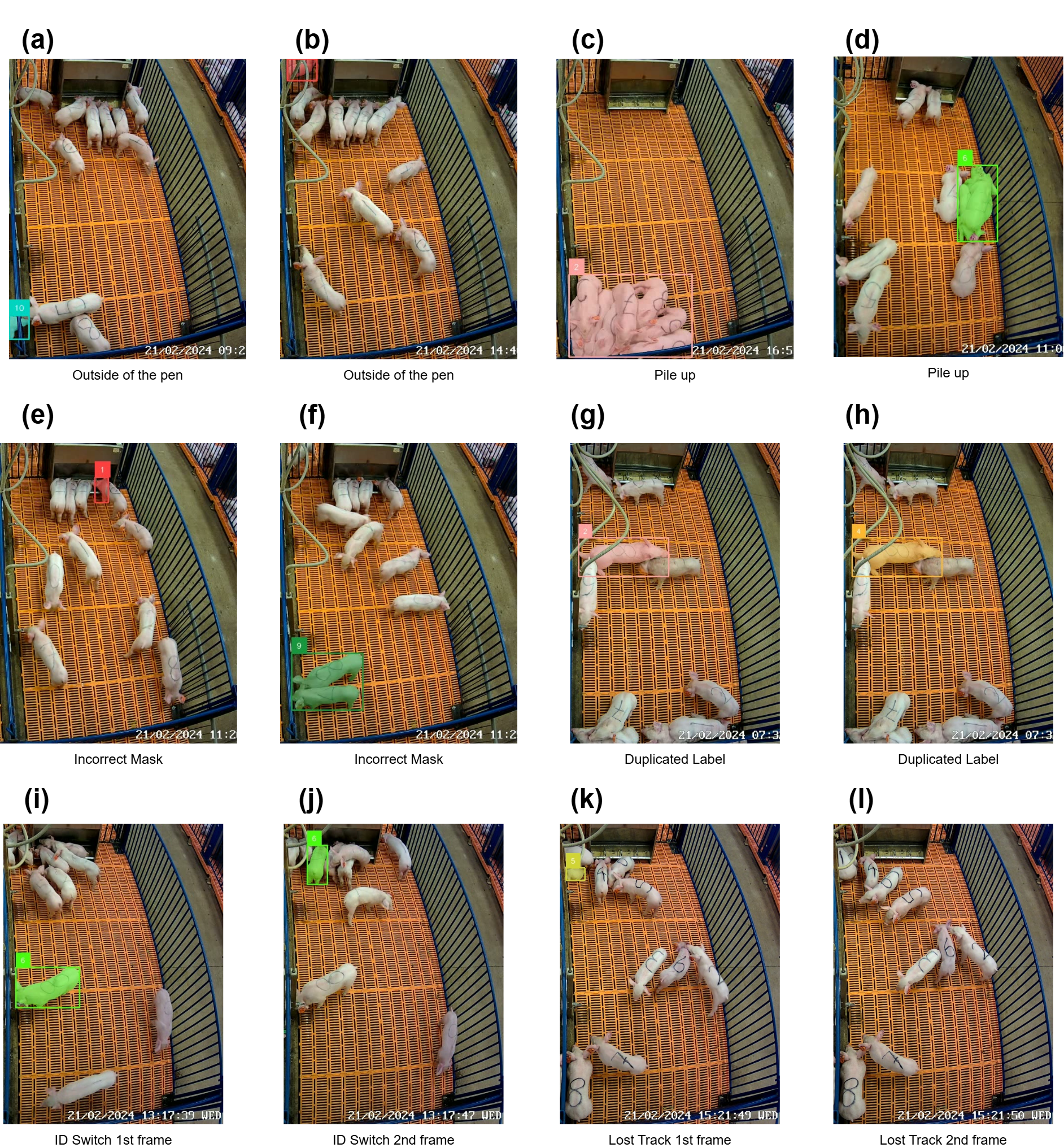}
    \caption{Short-term video segmentation examples by raw Grounded-SAM2. (a-b) Pigs outside of the pen. (c-d) Pigs piling up. (e-f) Incorrect masks. (g-h) Duplicated label for the same pig. (i-j) ID switch. (k-l) Lost track.}
\label{fig:shorterm_examples}
\end{figure}

Two primary challenges were identified: pigs located outside the target pen and pigs congregating in groups. As with animal detection (Section~\ref{sec:animal_detection}), the FM detected all visible pigs in the frame, including those in adjacent pens. This issue was addressed using a straightforward strategy: defining a target-tracking area and retaining only detections and masks within that region (Fig.~\ref{fig:shorterm_examples}a,b).

\begin{figure}[H]
    \centering
    \includegraphics[width=1\textwidth]{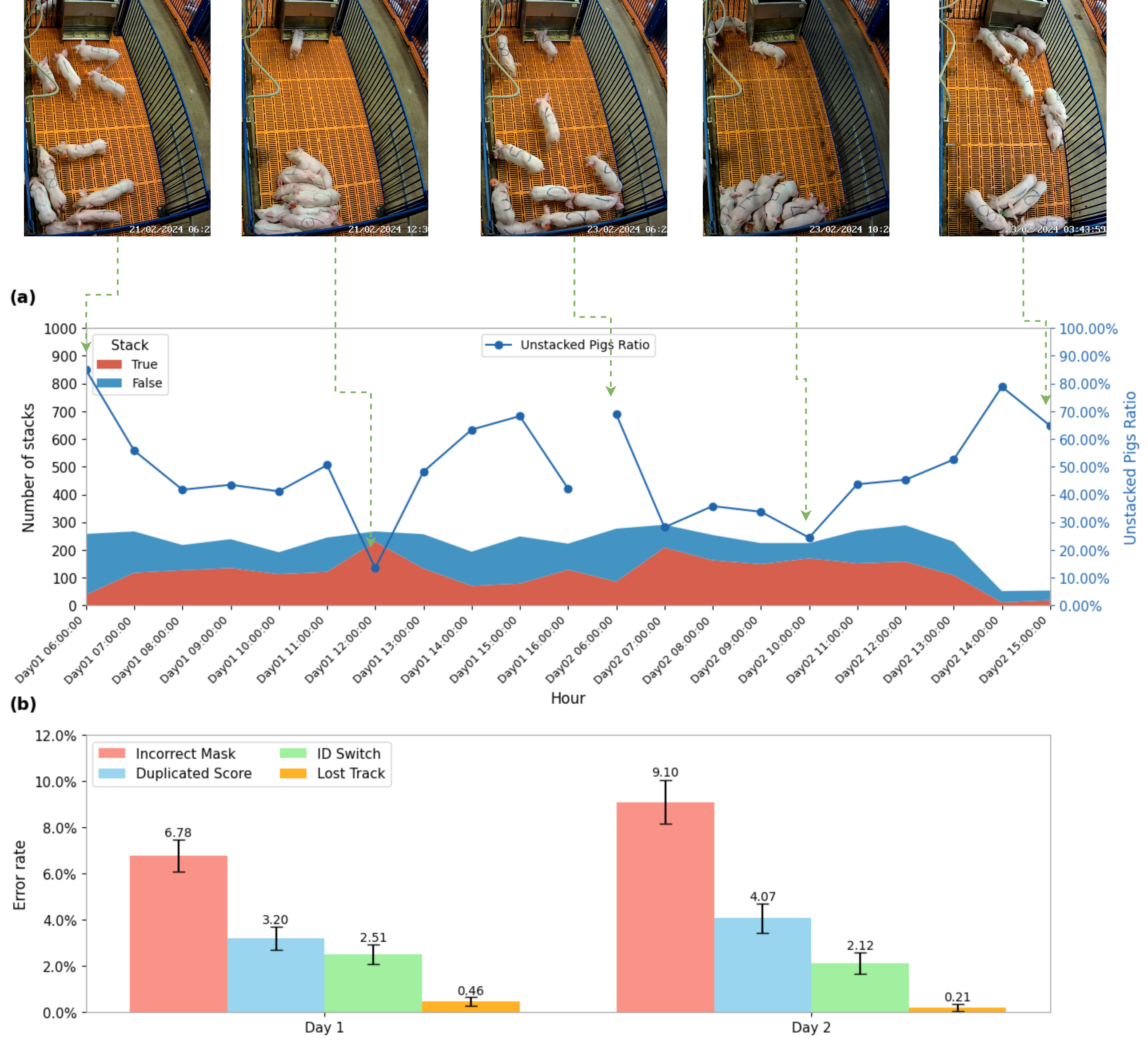}
    \caption{Short-term video segmentation results. (a) Number of stacked and unstacked pigs and the unstacked pig ratio across days and hours. and (b) Tracking errors within active pig tracks by raw Grounded-SMA2.}
\label{fig:shorterm_results}
\end{figure}

The second challenge arose when pigs slept in piles, forming large clusters in which individual body contours were no longer continuous or distinguishable. In such cases, the FM frequently grouped several pigs together (Fig.~\ref{fig:shorterm_examples} (c,d)). Nonetheless, pigs on the surface of these clusters could still be segmented accurately. Figure~\ref{fig:shorterm_results} (a) illustrates the distribution of pig-stacking status across hours and days, which reflects animal activity patterns: pigs were more active in the early morning when meals were provided (around 6 a.m.), became less active after feeding, and resumed activity later in the afternoon. As pigs piled up, occlusion became severe once animals were covered by others. Current computer vision methods for tracking and segmentation cannot effectively overcome such occlusions, as algorithms cannot ``see through” hidden regions. Consequently, occluded pigs were frequently undetected, lost, or assigned incorrect identities. To mitigate these issues, in Section~\ref{sec:long-seg} we introduce three corrective modules—mask refinement, matcher, and re-identification—that reduce the adverse effects of stacking and piling.

Four tracking errors observed in auto-annotated videos were noted and recorded, and the results are presented in Fig.~\ref{fig:shorterm_results} (b). Tracking examples are shown in Fig.~\ref{fig:shorterm_examples} (e-l). Completely correct tracks were observed in 87.38\% ± 0.91\% of videos on the first day and 83.74\% ± 1.20\% on the second day. Error rates on the first day were 6.78\% ± 0.69\% (incorrect mask), 3.20\% ± 0.69\% (duplicated label), 2.51\% ± 0.41\% (ID switch), and 0.46\% ± 0.18\% (lost track). On the second day, errors increased slightly: 9.10\% ± 0.93\%, 4.07\% ± 0.90\%, 2.12\% ± 0.46\%, and 0.21\% ± 0.15\%, respectively. 

The most frequent issues were incorrect masks, followed by duplicated labels and ID switches, typically arising during high activity near feeders or due to occlusion. 

Incorrect masks often appeared as those that either missed parts of the pig’s body or incorrectly extended into neighboring pigs or the background, as shown in Fig.\ref{fig:shorterm_examples} (c-f). These errors were especially noticeable in frames with clustered or partially overlapping pigs, which made their boundaries ambiguous. In some cases, the problem persisted for several consecutive frames until a clearer view was available, at which point the mask returned to the correct contour. While these errors raised the short-term error rate, their influence on overall tracking reliability was limited. In other cases, however, the problem persisted and remained unresolved within the same video clip. Importantly, in the long-term segmentation pipeline (section~\ref{sec:long-seg}), such persistent errors were largely avoided by carrying forward consistent masks across successive short-term video clips. This design eliminated the need to restart tracking and segmentation from scratch; instead, the system built on previously correct masks, which effectively suppressed the influence of local mask failures. 

For duplicated label errors shown in Fig.~\ref{fig:shorterm_examples} (g,h), after incorporating the mask-level non-maximum suppression procedure, these errors were effectively eliminated on both days. The refinement step ensured that only a single, highest-confidence mask was retained in overlapping regions. 

By substituting SAM2-Tiny with SAM2-Large and correcting errors with bidirectional SAM2 initialized from a valid anchor frame, segmentation stability was notably enhanced, and most ID switch and lost-track issues (Fig.~\ref{fig:shorterm_examples}(i-l)) were resolved. Specifically, 30 out of 33 problematic cases on the first day and 13 out of 15 cases on the second day were successfully corrected. The remaining four tracks still contained ID switch errors, mainly because heavy occlusion prevented us from locating a clear anchor frame with ten valid detections in the 1-minute video clips. 

\subsection{Long-Term Video Segmentation}
\begin{figure}[H]
    \centering
    \includegraphics[width=1\textwidth]{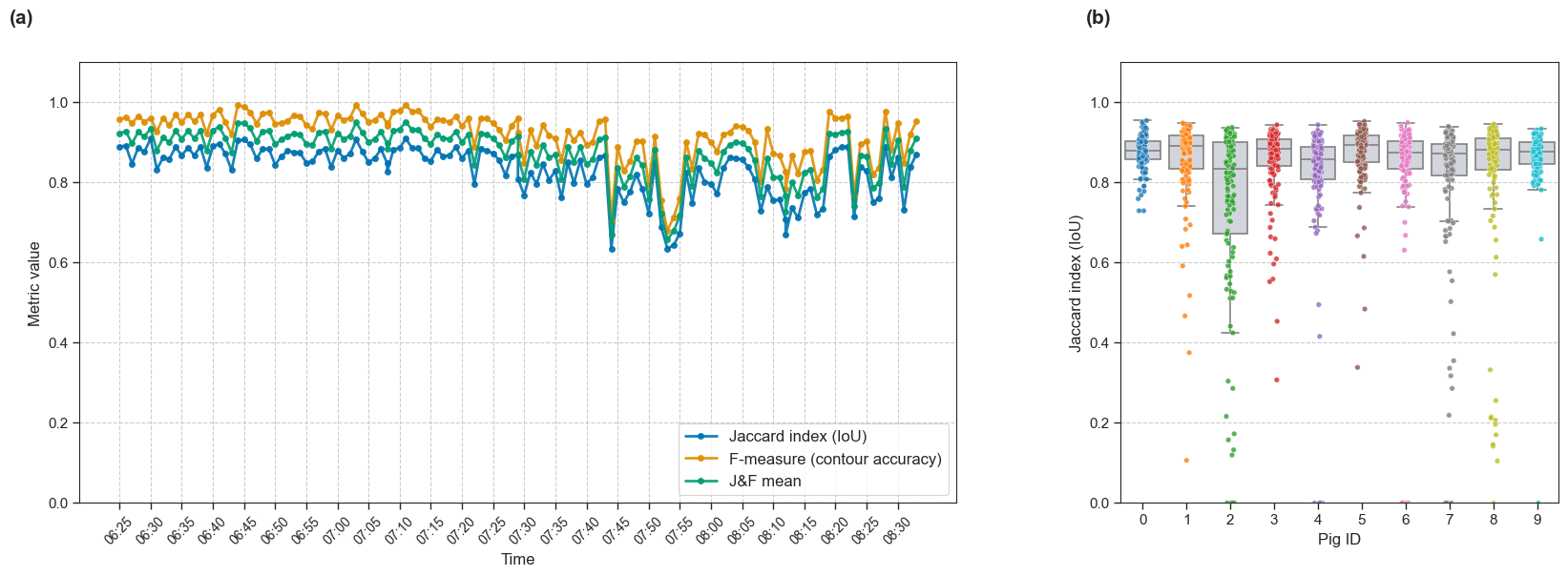}
    \caption{
    Quantitative results for long-term video segmentation. Modified foundation model performance using the $J\&F$ metric. (a) Mean $J$, $F$, and $J\&F$ scores over the time series. (b) $J$ scores for each individual pig.}
\label{fig:long_term_results}
\end{figure}

Long-term video segmentation was performed on six 22-minute-long videos, split into 1-minute video clips. Our modular pipeline maintained consistent pig identities across 1-minute video clips using six key modules: initializer, tracker, matcher, mask refinement, re-identification, and post-quality control. As shown in Fig.~\ref{fig:long_term_results}, the quantitative evaluation yielded region similarity ($\mathcal{J}$) of 0.83, contour accuracy ($\mathcal{F}$) of 0.92, and $\mathcal{J}\&\mathcal{F}$ mean of 0.87, consistent with high-quality segmentation, alongside favorable tracking metrics, MOTA of 0.99, and MOTP of 90.70\%. Among 1306 ground truth masks, there are 1293 true positives, 13 false negatives, 0 false positives, and 0 id switches.

Figure~\ref{fig:long_term_results}(a) further illustrates the change of segmentation quality over time, where $\mathcal{J}$, $\mathcal{F}$, and their average remain consistently high across time, with only occasional drops after index 80 (around 07:30 am) that coincide with challenges shown in Fig~\ref{fig:shorterm_results}(a), the pig activities dropping down after early morning feeding. 

Figure~\ref {fig:long_term_results} (b) breaks down the performance at the individual pig level. At the individual level, the average $\mathcal{J}$ values ranged from 0.74~$\pm$~0.24 (ID2) to 0.88~$\pm$~0.04 (ID0). Most pigs maintained $\mathcal{J}$ scores above 0.80, indicating consistently strong segmentation and contour accuracy across individuals. Larger standard deviations for some pigs (e.g., ID2, ID7, ID8) highlight challenging cases with more frequent occlusion or boundary interactions (as shown in the supplementary Fig.1), but overall performance remained stable across all tracked animals, with no identity switches observed. Together, these results confirm that GSAM2 maintains robust long-term segmentation and tracking accuracy at both the group and individual levels, making it well-suited for extended monitoring of group-housed pigs.

\begin{figure}[H]
    \centering
    \includegraphics[width=1\textwidth]{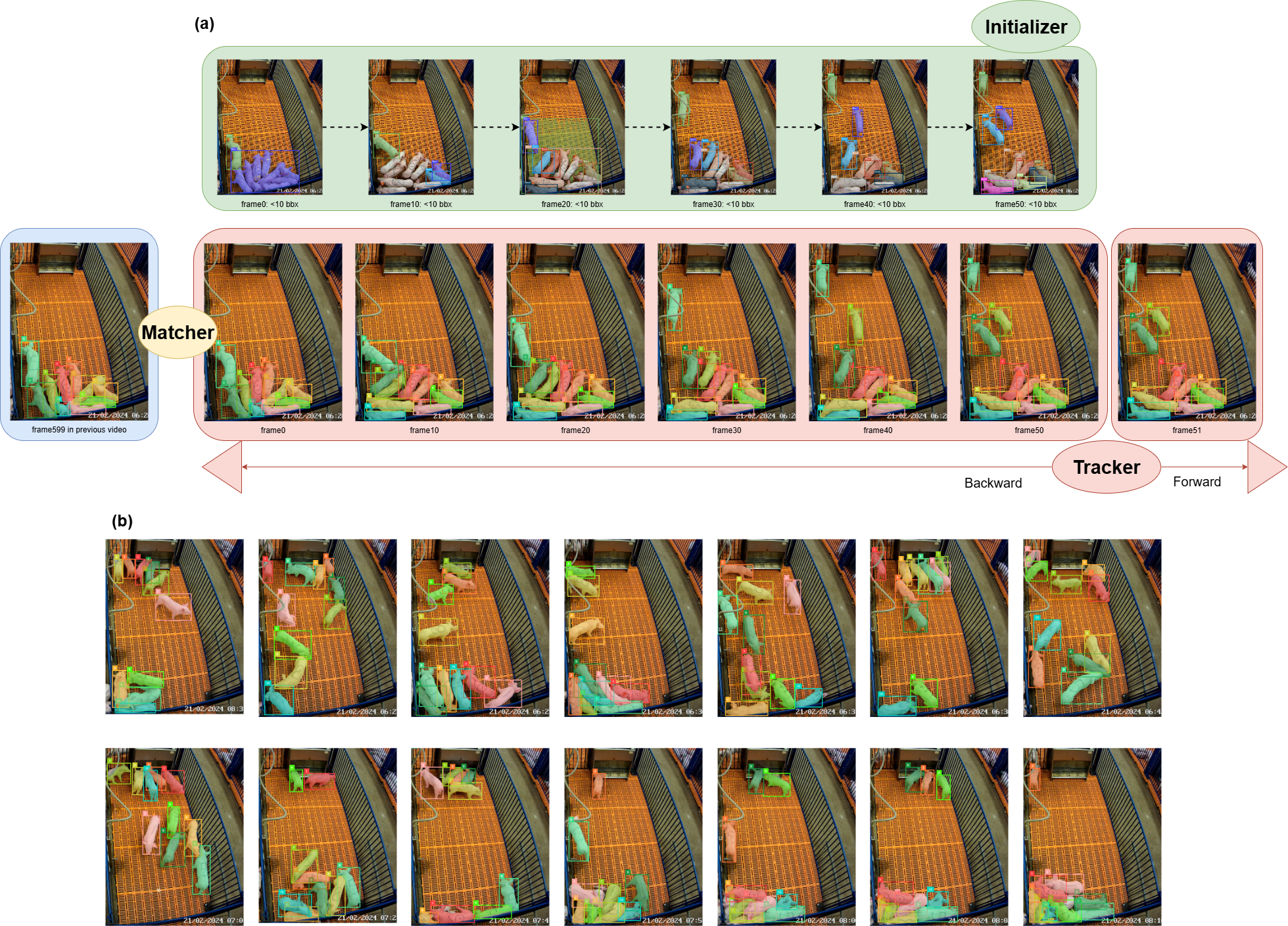}
    \caption{Qualitative results for long-term video segmentation. (a) The \textit{Initializer} selects an anchor frame, the \textit{Tracker} propagates masks forward and backward, and the \textit{Matcher} ensures identity consistency across video clips. (b) More examples from other time points.}
\label{fig:initializer}
\end{figure}

As shown in Fig.~\ref{fig:initializer}, the qualitative examples demonstrate that the modified FM achieves extended-duration video segmentation. The process began with the Initializer (green box), which selected the most suitable anchor frame within each video clip; in the example, frame 50 was chosen. Starting from this anchor, the Tracker (red circle) propagated segmentation masks and pig identities forward and backward across the clip, enabling robust coverage despite motion and occlusion. Finally, the Matcher module (yellow circle) aligned the first frame of the current clip (red box) with the last frame of the previous clip (blue box), which ensured consistent pig identities across consecutive video segments. This design effectively reduced identity switches and segmentation drift. Figure~\ref{fig:initializer}(b) provides further examples from periods of varying activity, including high activity, low activity, feeding, drinking, and resting, confirming that the model consistently improved temporal continuity and tracking reliability across a wide range of behaviors. 

\begin{figure}[H]
    \centering
    \includegraphics[width=1\textwidth]{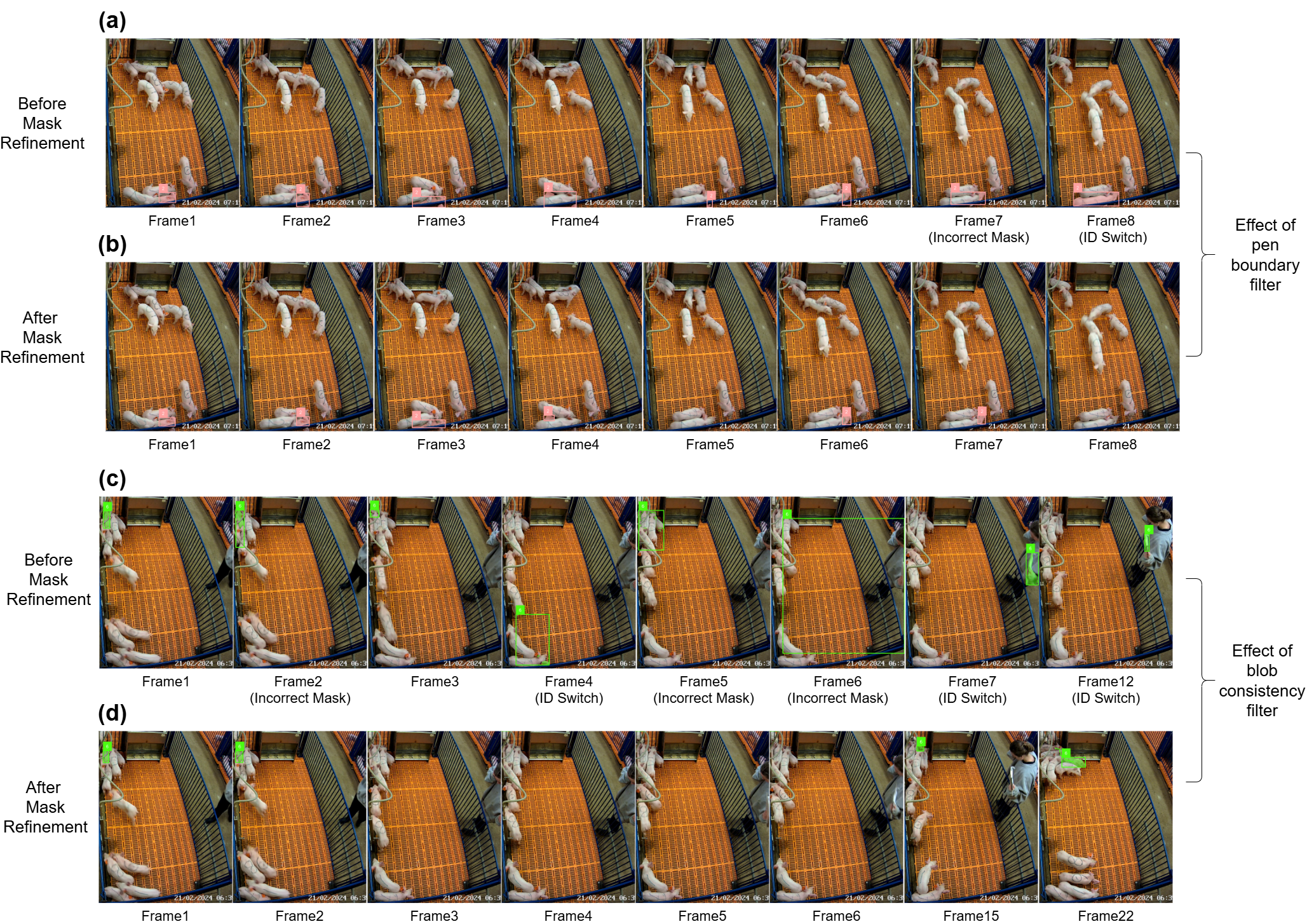}
    \caption{Qualitative results for long-term video segmentation. The \textit{mask refinement} module corrects errors by preventing masks from drifting outside the pen (pen boundary filter, shown in (a) and (b)) and from jumping or switching identities across frames (blob consistency filter, shown in (c) and (d)).}
\label{fig:mask_refine}
\end{figure}

As shown in Fig.~\ref{fig:mask_refine}, the mask refinement and post–quality control modules effectively corrected errors, removed duplicate masks, and eliminated persistent tracking failures. In panel (a), when pigs were stacked and one individual was completely covered, the mask frequently drifted from the pig’s body to a small visible region (an ear) outside the pen, and then switched to another pig, resulting in ID errors. This behavior reflected the FM’s difficulty in distinguishing closely packed pigs from small partial features. With the addition of the boundary filter (panel b), such erroneous masks were prevented: the filter suppressed jumps to regions outside the pen and temporarily removed a pig’s mask when the body was fully occluded, restoring it only once the pig reappeared. Panel (c) illustrates a more challenging case, where multiple pigs crowded into the corner of the pen following human intervention. Without refinement, the model produced incorrect masks spanning multiple pigs (frames 2, 5, and 6) or switched identities across pigs (frames 4, 7, and 12). By incorporating the blob consistency filter, masks were constrained to remain near their main blobs, avoiding large spatial jumps, and were appropriately cleared when the pig’s body disappeared, then reassigned once the pig became visible again.

Across all tasks, recurring challenges included heavy occlusion, prolonged inactivity, and visual similarity among pigs. While many errors were mitigated through rule-based filtering and modular post-processing, certain error types, particularly those arising from ambiguous detections in crowded frames, would likely require additional annotated data and targeted model training for further improvement. 

\section{Discussion}
This study demonstrates, for the first time to our knowledge, the feasibility of a fully foundation-model (FM) based, zero-shot framework for long-term video monitoring of group-housed pigs. Building on the core FM backbone, we further developed a modular workflow, including initializer, matcher, mask-refinement, re-identification, and post-quality-control modules, to achieve identity-consistent tracking across multi-hour video sequences. Together, these components enable label-free animal detection, segmentation, and identity maintenance under farm conditions that involve mid-dense animal populations. 

\subsection{Conceptual significance of foundation model-based framework}
Traditional supervised learning pipelines for livestock monitoring require extensive labeled datasets and frequent retraining for each new farm environment, limiting scalability~\citep{kamilaris2018deep, alameer2020automated}. In contrast, our FM-based framework reuses a single large pretrained model and requires labels only for early stage model development and evaluation. This shift away from training-intensive SLMs enables a far more scalable workflow while still transferring effectively to the visually complex setting of group-housed pigs~\citep{bommasani2021opportunities}.

We aim to redefine the analytical workflow in digital phenotyping by leveraging FM and zero shot learning. Several key differences distinguish FM-based pipelines from SLM-based approaches. First, ground-truth labels are required only for post-hoc evaluation, and the amount of annotation needed is substantially lower than what is necessary for training SLMs. For instance, in our long-term video segmentation pipeline, we manually annotated only one frame per minute video; once the system stabilizes for a given camera–pen configuration, this effort can be further reduced to one frame every 20 minutes or more. Second, because FM do not require domain-specific training, the framework eliminates the substantial computational burden, hyperparameter tuning, and architectural engineering associated with SLMs development. Third, evaluation becomes more streamlined and resource-efficient. Alongside standard metrics computed on sampled frames, such as MOTA, MOTP, and J\&F during model development, we perform a user friendly procedure: the tracking pipeline automatically isolates the frames with potential errors (Fig.~\ref{fig:long_term_results}), and a trained observer reviews only these flagged segments to validate or correct mask and identity assignments, making the process practical even without powerful computing infrastructure. Fourth, FM-based pipelines offer broader validity and more flexible deployment scenarios. Whereas classical cross-validation requires splitting the data into training, validation, and testing sets, FM-based approach can evaluate directly on held-out test data, and in principle, the entire dataset can be treated as a test set without sacrificing any portion for training. The pipeline can operate even on a single image, enabling scalable deployment across farms and environments.

In addition, FM offer powerful tools for dataset augmentation, reducing reliance on large manually annotated datasets and supplying high-quality synthetic or semi-automated labels when supervised models are needed. FM can accelerate the annotation process by assisting with mask generation from raw farm images~\cite{li2024promote}. For example, annotators can provide minimal prompts—such as points or bounding boxes—within tools like LabelMe~\citep{russell2008labelme}, CVAT~\citep{cvat2020}, or Roboflow~\citep{roboflow2022} to obtain high-quality segmentations, or automatically segment all objects in a frame using SAM series~\citep{kirillov2023segment}. These strategies substantially reduce the need to draw mask polygons manually. Although the data augmentation is not the study's focus, our study also generated several datasets with open access.
 
\subsection{Performance interpretation}
Quantitative and qualitative results highlight the robustness of the FM-based pipeline across both segmentation and tracking tasks. For long-term segmentation, region similarity ($J = 0.83$), contour accuracy ($F = 0.92$), and the combined $J\&F = 0.87$ demonstrate close alignment with ground truth. Supervised models for group-housed pigs typically report IoU values of 0.80–0.88 for static images~\citep{van2021individual, huang2023semi, liao2025yolov8a}. Our video-based IoU ($J$) of 0.83 for 18-day-old nursery pigs falls within this range, indicating that an FM-centered workflow can achieve segmentation accuracy comparable to supervised methods, even under more complex temporal conditions.

Tracking performance shows similarly strong results. With a MOTA of 0.99 and MOTP of 90.70\%, pig identities remained exceptionally stable across two-hour sequences. This reliability reflects not only the short-term continuity provided by SAM2 but also the contribution of our quality-control workflow, which automatically flags questionable frames for targeted review. Prior supervised systems report MOTA values of 0.90–0.98 and MOTP values of 80–90\% across a range of durations~\citep{t2020long, guo2023enhanced, odo2025re, huang2025behavior}, reflecting a diversity of approaches in which some methods incorporate complementary identity cues, such as ear-tag recognition, to support extended tracking periods. Even so, our zero-shot long-term results fall well within the performance range of these supervised approaches.

A key factor enabling this performance is our shift away from Kalman filter–based motion prediction. Trackers such as SORT and DeepSORT depend on linear motion assumptions that are often violated in group-housed pig environments, where movements are irregular and frequently interrupted by occlusion and stacking. Recent MOT studies have underscored the limitations of Kalman-based models under such conditions~\citep{cao2023observation}. Because SAM2 already provides stable short-term temporal propagation, explicit motion models are unnecessary. Instead, long-term identity maintenance relies on robust cross-segment association: frozen SAM2 embeddings and spatial–temporal proximity are integrated into a unified cost matrix and optimized using the Hungarian algorithm~\citep{kuhn1955hungarian}. This contrasts with SORT~\citep{Bewley2016SORT}, DeepSORT~\citep{Wojke2017DeepSORT}, and the OC-SORT family~\citep{Cao2023OCSORT, Maggiolino2023DeepOCSORT}, achieving comparable stability while preserving the zero-shot nature of an FM-driven pipeline.

Short-term tracking results further illustrate the effectiveness of this modular design. Across more than 4,900 auto-annotated clips, over 85\% exhibited fully correct tracks. The remaining issues, such as incorrect masks, duplicated labels, and occasional ID switches, were concentrated in periods of high activity or heavy occlusion. Once bidirectional tracking and mask-refinement modules were incorporated, most of these issues were resolved automatically, underscoring the value of combining FM inference with targeted post-processing.

Beyond comparisons with supervised methods, our results also advance the state of FM-based tracking itself. Existing FM-assisted livestock tracking systems typically operate on short video segments or rely on supervised initialization to stabilize identity tracking~\citep{yang2024innovative, yang2025computer, bibinbe2025multi}. By contrast, our pipeline maintains identity consistency across continuous multi-hour videos without any fine-tuning or annotated training data. This demonstrates that FM-centered workflows, when augmented with carefully designed modular components, can support fully automated, long-term behavioral monitoring in commercial pig production environments.

\subsection{Sources of error and remaining challenges}
Despite these promising results, several persistent challenges were observed. Occlusion and stacking remain the most significant obstacles~\citep{wurtz2019recording}. When pigs rest in physical contact, their body contours merge, preventing the algorithm from isolating individual masks because visual evidence of boundaries disappears. Although the blob-consistency and boundary-filter modules substantially reduced erroneous mask jumps and duplicate IDs, complete recovery of fully hidden pigs remains difficult. 

Night-time infrared (IR) footage presented a second major challenge. Grayscale IR images lack color and fine-texture cues, reducing recall and contour precision compared with daytime RGB footage~\citep{henrich2026benchmarking}. SAM2 maintained temporal continuity for visible pigs, but overall segmentation quality decreased under low illumination. Pre-processing techniques such as adaptive contrast enhancement, denoising, or the recently proposed “Depth-Oriented Gray Image” transformation~\citep{seo2025depth} could help restore structural information.

Another subtle source of discrepancy arises from differences between the model’s default interpretation and dataset-specific labeling policies~\citep{northcutt2021pervasive}. FM often identified partially visible or outside-pen pigs as positives, whereas human annotators labeled them as negatives according to predefined inclusion rules. These inconsistencies do not necessarily reflect model failure but rather conceptual misalignment between universal model perception and localized annotation standards. For operational use, defining consistent annotation conventions that align with FM perception may improve evaluation fairness.

\subsection{Current limitations and future directions}
Although the FM-based pipeline demonstrates strong performance, several limitations highlight opportunities for future development. The present study was conducted in a single on-campus nursery facility, and broader external validity will require evaluating the system across diverse farms with varying housing designs, stocking densities, and management practices. Prior work has shown that cross-site variability can significantly affect livestock computer vision models~\citep{li2024piglife, jiang2023domain}, underscoring the importance of expanding the dataset and exploring domain adaptation or domain-generalization strategies.

Another limitation concerns the computational and memory demands of FMs. Architectures such as SAM2 are resource-intensive and are not yet practical for deployment on edge devices likely to be found in barns. Recent developments in lightweight segmentation and tracking models—including MobileSAM~\citep{zhang2023mobilesam}, MobileSAMv2, and Track-Anything frameworks~\citep{yang2023trackanything} offer promising avenues for improving efficiency. Knowledge distillation~\citep{hinton2015distillation, gou2021knowledge} represents an additional strategy for deriving compact student models that retain the capabilities of large FM while consuming far fewer computational resources. Developing such distilled models tailored to pig barn environments would significantly enhance deployability.

Identity recognition is another area that could be improved. Although our segmentation-aware association method performs well without a trained Re-ID network, extreme occlusion, dense clustering, and highly homogeneous appearance across nursery pigs can still lead to identity drift. Incorporating more advanced Re-ID mechanisms, such as transformer-based appearance encoders~\citep{he2021transreid}, self-supervised Re-ID approaches~\citep{yang2022selfreid}, or lightweight animal-specific descriptors, may further stabilize long-term identity tracking.

Finally, the system must contend with distribution shifts and the inherent semantic ambiguity in real-world barn environments. Grounding-based detectors occasionally fail to recognize all pigs when visual conditions degrade, when animals overlap, or when boundary cues between individuals become indistinct. Such challenges mirror broader concerns about the FM's sensitivity to domain shift~\citep{bommasani2021opportunities}. Emerging multimodal architectures offer promising avenues to address these limitations. SAM3, for example, improves text-prompt grounding and open-vocabulary segmentation capabilities, enabling more precise language-driven region identification in complex scenes. Sa2VA integrates SAM2 with LLaVA to achieve dense grounded visual understanding across both images and video~\citep{carion2025sam, yuan2025sa2va}. These developments suggest that future FM-based pipelines may benefit from richer text–vision interactions and more robust semantic grounding.

\section{Conclusions}
This study demonstrates that FM alone can enable reliable, label-free, and scalable long-term monitoring of group-housed pigs. Using a unified Grounded-SAM2 pipeline, we jointly achieved pig detection, instance segmentation, and multi-object tracking across both static images and multi-hour video sequences, fulfilling our objectives of evaluating FM performance across temporal scales, developing an integrated FM workflow, and identifying practical challenges for real-farm deployment. The strong segmentation accuracy and stable identity tracking achieved in a zero-shot setting provide clear evidence that FM-based systems offer a viable path toward autonomous precision livestock farming while highlighting opportunities for further improvement in handling occlusion, stacking, and low-light conditions.

\newpage
\section{Author contribution statement}
Ye Bi: Conceptualisation; Methodology; Formal analysis, Investigation; Software; Data curation; Visualisation; Writing-original draft; Writing-review and editing.
Bimala Acharya: Data curation; Resources; Investigation; Writing-review and editing
David Rosero: Data curation; Resources; Funding acquisition; Writing-review and editing.
Juan Steibel: Conceptualization; Methodology; Funding acquisition, Project administration, Supervision, Writing-original draft, Writing-review and editing. 

\section{Dataset availability}
Twenty video clips with 12,000 frames and their corresponding annotated JSON files, including both 100 standard tracks and 111 challenging tracks, have been uploaded to the OSF website(\url{https://osf.io/63ndh/}).


\section{Funding}
This work was supported by the Iowa Pork Producers Association [grant number: 24-103] and the National Institute of Food and Agriculture [Federal Award: 20256701544741]

\section{Conflict of interest}
The authors declare that there is no conflict of interest.

\section{Declaration of generative AI use}
Statement: During the preparation of this work, the author(s) used ChatGPT (OpenAI-ChatGPT5) to improve the readability and clarity of the English manuscript. After using this tool, the author(s) reviewed and edited the content as needed and take(s) full responsibility for the content of the published article.

\clearpage
\newpage 
\bibliographystyle{apalike} 
\bibliography{main}

\newpage

\section*{Tables}

\begin{table}[h]
\centering
\caption{Tracking errors and video challenges during short-term video segmentation evaluation}
\label{tab:track_error}
\begin{tabular}{p{4.5cm} p{9cm}}
\toprule
\textbf{Category} & \textbf{Description} \\
\midrule
\textbf{Incorrect mask} & Segmentation mask does not match the pig’s shape or location. \\
\textbf{Duplicated label} & The same pig is detected and tracked multiple times in a single frame. \\
\textbf{ID switch} & Identities are mistakenly swapped between pigs during tracking. \\
\textbf{Lost track} & Pig disappears from the mask sequence before the end of the video. \\
\hline
\textbf{Pigs outside of pen} & Pigs are detected outside the intended pen boundary. \\
\textbf{Pigs stacking up} & Two or more pigs overlap significantly, causing occlusion. \\
\bottomrule
\end{tabular}
\end{table}


\newpage
\section*{Figures}

\newpage

\newpage

\newpage

\newpage

\newpage

\newpage

\newpage

\newpage

\newpage

\newpage

\end{document}


\title{Automated Segmentation and Tracking of Group Housed Pigs Using Foundation Models}

\author[1]{Ye Bi}
\author[1,2]{Bimala Acharya}
\author[1]{David Rosero}
\author[1, *]{Juan Steibel}
\affil[1]{Department of Animal Science, Iowa State University, Ames, IA, 50010, USA}
\affil[2]{Interdepartmental Bioinformatics and Computational Biology, Iowa State University, Ames, IA, 50010, USA}

\date{}

\maketitle
\noindent 
$^{*}$ Corresponding author \\

\newpage

\section*{Pseudo-code}
\begin{algorithm}[H]
\caption{Mask Refinement Procedure}
\label{alg:mask_refine}
\begin{algorithmic}[1]
\Require 
    \begin{itemize}
        \item Mask for current frame
        \item \texttt{MIN\_AREA\_RATIO} = 0.30
        \item \texttt{MAX\_DISTANCE\_RATIOS} = [0.5, 1.5]
        \item \texttt{MAX\_PREV\_DIST} = 200 pixels
    \end{itemize}
\State Apply pen-boundary filter to restrict the mask to the valid enclosure
\State Extract blobs: compute areas $A_i$ and centroids $\mu_i$
\If{$\mu_{\text{prev}}$ exists}
    \State Select main blob closest to $\mu_{\text{prev}}$
\Else
    \State Select blob with largest area
\EndIf
\State Set $A_{\text{main}} \gets$ area of main blob, and compute $R \gets \sqrt{A_{\text{main}} / \pi}$
\State Compute thresholds:
    \begin{itemize}
        \item $A_{\text{min}} \gets$ \texttt{MIN\_AREA\_RATIO} $\times A_{\text{main}}$
        \item \textbf{If} $A_{\text{main}} \geq 1000$: $D_{\text{max}} \gets$ \texttt{MAX\_DISTANCE\_RATIOS[0]} $\times R$
        \item \textbf{Else}: $D_{\text{max}} \gets$ \texttt{MAX\_DISTANCE\_RATIOS[1]} $\times R$
        \item $D_{\text{prev}}^{\text{max}} \gets$ \texttt{MAX\_PREV\_DIST}
    \end{itemize}
\State Initialize cleaned mask as empty
\For{each blob $i$}
    \State Compute contour distance $d_{\text{main}}(i)$ to main blob
    \State Compute distance $d_{\text{prev}}(i)$ to previous centroid
    \If{$i$ is main blob}
        \State Keep blob $i$
    \ElsIf{$A_i \geq A_{\text{min}}$ and $d_{\text{main}}(i) \leq D_{\text{max}}$ and $d_{\text{prev}}(i) \leq D_{\text{prev}}^{\text{max}}$}
        \State Keep blob $i$
    \EndIf
\EndFor
\If{no blobs kept}
    \If{frame is first}
        \State Use original mask
    \Else
        \State Set mask to empty
    \EndIf
\EndIf
\If{mask is not empty}
    \State Update $\mu_{\text{prev}}$ using image moments
\EndIf
\State \Return cleaned mask
\end{algorithmic}
\end{algorithm}

\newpage
\begin{algorithm}[H]
\caption{Feature- and Location-Aware Re-Identification (ReID)}
\label{alg:reid}
\begin{algorithmic}[1]
\Require New image $I^\text{new}$, old image $I^\text{old}$, new masks $\{\mathcal{M}_i^\text{new}\}_{i=1}^{N_\text{new}}$, old masks $\{\mathcal{M}_j^\text{old}\}_{j=1}^{N_\text{old}}$
\Ensure Identity mapping from new to old objects
\vspace{0.5em}
\State \textbf{Enhance and Crop:} For each mask in $I^\text{new}$ and $I^\text{old}$:
\begin{itemize}
    \item Crop the masked region
    \item Apply gamma correction and histogram equalization
\end{itemize}
\State \textbf{Feature Extraction:} Pass enhanced crops through SAM2 encoder to extract feature vectors $\mathbf{f}_i^\text{new}$ and $\mathbf{f}_j^\text{old}$ via global average pooling
\State \textbf{Similarity Computation:} For each pair $(i,j)$:
\begin{itemize}
    \item Compute cosine similarity: 
    \begin{align}
    S^{\text{cos}}_{i,j} &= \frac{\mathbf{f}_i^{\text{new}} \cdot \mathbf{f}_j^{\text{old}}}{\|\mathbf{f}_i^{\text{new}}\| \, \|\mathbf{f}_j^{\text{old}}\|}
    \end{align}
    \item Compute IoU between masks: 
    \begin{align}
    S^{\text{IoU}}_{i,j} &= \frac{|\mathcal{M}_i^{\text{new}} \cap \mathcal{M}_j^{\text{old}}|}{|\mathcal{M}_i^{\text{new}} \cup \mathcal{M}_j^{\text{old}}|} 
    \end{align}
    \item Compute centroid distance and normalize: 
    \begin{align}
    S^{\text{centroid}}_{i,j} &= 1 - \frac{\|\mathbf{c}_i^{\text{new}} - \mathbf{c}_j^{\text{old}}\|_2}{d_{\text{max}}}  
    \end{align}
\end{itemize}
\State \textbf{Cost Matrix:} Combine the similarities into a weighted cost:
\[
\text{Cost}_{i,j} = 1 - \left( \alpha \cdot S^{\text{cos}}_{i,j} + \beta \cdot S^{\text{IoU}}_{i,j} + \gamma \cdot S^{\text{centroid}}_{i,j} \right)
\]
\State \textbf{Identity Assignment:} Apply the Hungarian algorithm to the cost matrix to determine optimal identity mapping
\State \textbf{Update Tracker:} Reorder object identities in the new frame to match the assignment result
\end{algorithmic}
\end{algorithm}
